# A Stacking Ensemble Approach for Supervised Video Summarization

Yubo An, Shenghui Zhao and Guoqiang Zhang, *Member, IEEE*

*Abstract*—Video summarization methods are usually classified into shot-level or frame-level methods, which are individually used in a general way. This paper investigates the underlying complementarity between the frame-level and shot-level methods, and a stacking ensemble approach is proposed for supervised video summarization. Firstly, we build up a stacking model to predict both the key frame probabilities and the temporal interest segments simultaneously. The two components are then combined via soft decision fusion to obtain the final scores of each frame in the video. A joint loss function is proposed for the model training. The ablation experimental results show that the proposed method outperforms both the two corresponding individual method. Furthermore, extensive experimental results on two benchmark datasets shows its superior performance in comparison with the state-of-the-art methods.

## I. INTRODUCTION

With the rapid development of the mobile networks, the self-media rise results in massive video data. Hence the computer vision technology that can efficiently browse, watch and summarize videos, referred to as video summarization, has attracted more and more attention [1]. Currently, the supervised learning methods for video summarization [2][3][4], which use the training data that is composed of the ground-truth labels manually generated with human preferences, are usually superior to the unsupervised methods [5][6][7][8] because they can implicitly learn human preferences.

The current deep learning based approaches for video summarization are approximately classified into shot-level and frame-level by the partition strategies. The frame-level deep learning methods usually rely on Long Short-Term Memory (LSTM) or attention mechanism to capture long-term and short-term dependencies within a video, and use appropriate frame scoring networks to predict the probability of each frame being selected into the video summary [1][3][9][10]. But the prediction scores of the video frames in the same semantic segment cannot accurately represent the importance of the corresponding segment without temporal consistency constraints. And the LSTM-based methods often suffer from low variation in prediction probabilities, which would have a restricted impact when generating the final summary [11].

To solve the problem of temporal consistency, a typical shot-level method is predicting the selection scores on the segmented shots rather than on each frame, which facilitates exploiting temporal similarities and dependencies within a video [12-15]. However, there are two disadvantages as below for the shot-level strategy: (1) the evaluation form is relatively simple and all the frames in same segment are given the same importance score, which results in a lack of diversity when generating summaries; (2) the importance score predicted by this method is not sufficiently accurate due to the errors caused by the prediction. And the unimportant frames in the segment may be given higher scores.

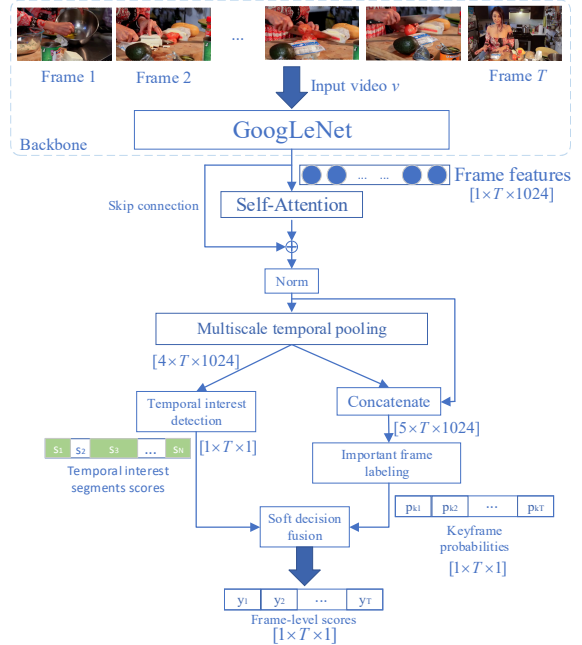

Fig. 1. The structure of the proposed approach

Thus, the shot-level and frame-level methods are regarded as two different methods and are usually used individually. In order to separate the subtask of temporal consistency problem from the prediction of the frame-level importance scores and refine the importance scores of each frame under the constraints of temporal interest segments, we propose an attention-based stacking ensemble approach for supervised video summarization to learn the underlying contact between the two methods based on our previous work [16]. Inspired by the temporal interest proposals strategy proposed by [14] and the sequence labeling formulation in [2], the proposed model first encodes the features via a self-attention mechanism, and then uses two predictors to predict the interest segments scores and the frame-level importance probabilities simultaneously as the intermediate features. Finally, the intermediate features are combined as the input to a soft decision fusion model to estimate the final scores of each frame in the video.

The main innovations and contributions of our video summarization method are as follow.

*1)* A stacking ensemble approach is proposed for supervised video summarization to learn the potential complementarity between segment and frame partition.

2) A joint loss function is proposed to train the stacking model with the interest proposal label, the important frame label and the important score label.

3) To the best of our knowledge, this work is the first attempt to fuse the frame-level and shot-level strategies for video summarization.

## II. APPROACH

The approach uses soft decision fusion to combine key frame probabilities and temporal interest segments. As shown in Fig. 1, given a video $v$ with $T$ frames. A pre-trained backbone network is used to extract the visual features for each video frame. After feature extraction, the video feature sequence is denoted as $X \in \mathbb{R}^{d \times T}$, where $d$ is the feature dimension. To capture the temporal long-range dependencies, we use self-attention mechanism [17] and skip connection to re-encode frame features to obtain the final representation. Temporal average pooling $1d$ is used in time dimension with the kernel size of 4, 8, 16, 32 and the stride of 1 to avoid temporal warping or cropping, which is called multiscale temporal pooling [14]. The processing of the pooled features is divided into two branches as below.

(1) The pooled features are fed into the temporal interest detection module to obtain the interest segments scores denoted as $P_S = [\rho_{1,s_1}, ..., \rho_{t,s_n}, ..., \rho_{T,s_N}] \in \mathbb{R}^T$, where $\rho_{t,s_n}$ is the interest score of the *t-th* frame dividing into the *n-th* segment in a total of $N$ segments. Thus, the video $v$ that consist of a sequence of consecutive frames is temporally divided into $N$ disjoint segments, and the frames in each segment are assigned the same interest score.

(2) The combination of the multiscale temporal pooling features and the final representation features are put into the important frame labeling model to predict the key frame probabilities of each frame in video as $P_K = [p_{k1}, ..., p_{kt}, ..., p_{kT}] \in \mathbb{R}^T$, where $p_{kt}$ is the *t-th* frame's probability of being selected as a key frame.

Finally, $P_S$ and $P_K$ are integrated into a soft decision fusion mechanism to predict the final frame-level scores as $Y = [y_1, ..., y_t, ..., y_T] \in \mathbb{R}^T$, where $y_t$, ranging from 0 to 1, is the *t-th* frame-level score of a video. The higher the score a frame obtains, the higher probability the frame will be selected into the final summary with.

### A. Feature Extraction

Following the previous methods [1][2][3], we uniformly down-sample the videos to 2 *fps*. Then we take the output of the *pool5* layer in the pretrained GoogLeNet [18] as the feature descriptor for each video frame. The dimensionality of the feature descriptor is 1024.

### B. Temporal interest detection

Inspired by [14], we directly adopt the temporal interest proposals generation strategy to generate temporal interest segments scores by pre-defined multi-scale intervals. E.g., at the *t-th* temporal location, $K$ interest proposals are appointed with the fixed range $[t - \lambda_k/2, t + \lambda_k/2)$, $k = 1, 2, ..., K$, where $\lambda_k$ is the duration of the *k-th* interest proposal. Therefore, $K \times T$ interest proposals are totally produced in a video sequence with $T$ frames. In the training stage, a proposal is positive when its temporal Intersection over Union (tIoU) with any ground truth proposal is higher than 0.6, or negative with $0 \leq \text{tIoU} < 0.3$. The proposals within $0.3 \leq \text{tIoU} < 0.6$ are discarded in loss calculation. $l_k$ is calculated by the conclusive equations in [14] as {4, 8, 16, 32} to cover all ground-truth proposals with the durations from 1 to 44.

The details of the model are shown in Fig.2 (a). *Fc-1* attempts to connect different scaled pooling layers and can effectively prevent the overfitting, and includes *tanh* and *layer-normalization*. There are two sibling output subbranches following *Fc-2*. The first outputs importance scores of proposals, and the second one outputs the associated center and the proposals length offsets.

Finally, min-max normalization is implemented on refined proposals by the non-maximum suppression (NMS) [19] to obtain the temporal interest segments scores.

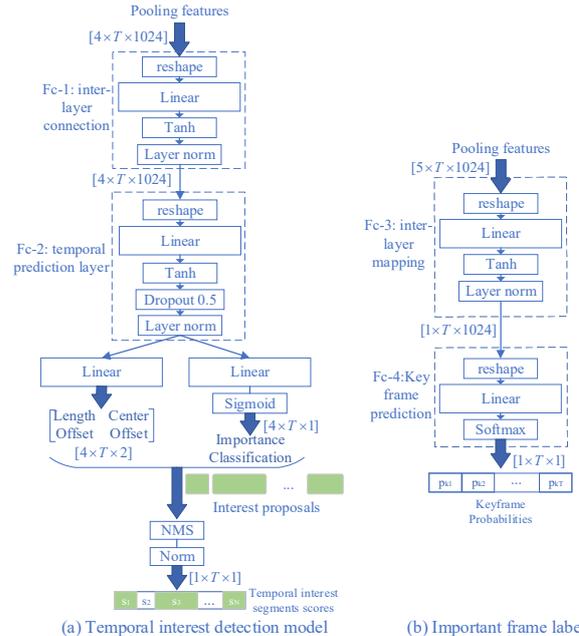

Fig. 2. The detailed components of the interest detection model and important frame labeling model

### C. Important frame labeling

Just like semantic segmentation indicating the semantic label of the corresponding pixels, important frame labeling model is formulated as a sequence labeling problem to indicate the semantic label of the corresponding frame in time dimension [2]. The details of the model are shown in Fig.2 (b). *Fc-3* attempts to map the multi-layer features into single layer. *Fc-4*, including linear layer and *softmax* function, attempts to predict the frame-level probabilities with the output dimension of $T \times C$, where $T$ is the number of frames, and $C$, the dimension of the output channel, is 2 since we need scores corresponding to 2 classes (keyframe or non-keyframe) for each frame.

*D. Soft decision fusion*

Temporal interest segments scores and frame-level probabilities are integrated as intermediate features into a soft decision fusion mechanism to predict the final frame-level scores. In this model, the most common method is to average the two scores with the same weight. To introduce more nonlinearity and achieve higher fusion performance, we use a simple multilayer perceptron as the meta-learner that utilizes the two intermediate features to fit final ground-truth frame-level scores.

*E. Learning*

Multi-task loss is proposed to train the model jointly. The objective function is obtained as

$$L = L_{cls} + L_{reg} + L_{pre} + L_{mse} \quad (1)$$

where $L_{cls}$ and $L_{reg}$, accounting for the temporal interest detection module, are the classification loss and the regression loss respectively, $L_{pre}$ is the prediction loss of the frame-level probabilities for important frame labeling, and $L_{mse}$ is the fitting loss of the meta-learner. The four losses are considered to be equally important.

Specifically, focal loss [20] is used to obtain more accurate detection of the interesting proposals as

$$L_{cls} = -\frac{1}{N}\sum_{m=1}^{N}(1-p_t(m))^{\gamma}\log(p_t(m)) \quad (2)$$

$$p_t = \frac{\exp(x[class])}{\sum_j \exp(x[j])} \quad (3)$$

where $M$ is the total number of predicted proposals (including positive and negative proposals), and $p_t(m)$ is the probability of the *n-th* proposal being classified into the corresponding ground-truth interest proposal labels, which is actually represented by the *softmax* function. $\gamma$ is a hyperparameter that is artificially set to be 1.

The regression loss $L_{reg}$ is actually the positioning regression for the positive interest proposal, and is obtained as

$$L_{reg} = \frac{1}{N_{pos}}\sum_{i=1}^{N_{pos}} p_t(i) \frac{1}{Q}\sum_{q=1}^{Q} smooth_{L_1}(t_i(q) - t_i^*(q)) \quad (4)$$

$$smooth_{L_1}(x) = \begin{cases} 0.5x^2 & if\ |x|<1, \\ |x|-0.5 & otherwise \end{cases} \quad (5)$$

where $N_{pos}$ is the number of positive interest proposals, $p_t(i)$ is the probability of the *i-th* positive interest proposals being classified into the corresponding ground-truth interest proposal labels, $t_i(t_i^*)$ is the predicted (ground-truth label) *i-th* group of positioning regression, and each group contains $Q$ parameters. The regression loss of positioning regression is calculated by the $smooth_{L_1}$ loss.

Specifically, the predicted location offset $t_i = (\delta c_i, \delta l_i)$ contains the center position and the length offsets between the generated proposals and the pre-defined proposals. The ground truth location offset $t_i^* = (\delta c_i^*, \delta l_i^*)$ is obtained as follows.

$$\delta c_i^* = (c_i^* - c_i)/l_i, \quad \delta l_i^* = \ln(l_i^*/l_i) \quad (6)$$

For the prediction loss of frame-level importance scores, the weighted focal loss is proposed because the categories of temporal important frames and non-important frames are extremely unbalanced, which is obtained as

$$L_{pre} = -\frac{1}{T}\sum_{i=1}^{T}\omega(i)(1-p_t(i))^{\gamma}\log(p_t(i)) \quad (7)$$

where $T$ is the total number of temporal frames, $p_t(i)$ is the probability of the *i-th* frame being classified into the corresponding ground-truth label as shown in Eq. (3), $\omega(i)$ is the category weight of the ground-truth classification ($\omega$) of the *i-th* frame, and $\omega = \frac{median\_freq}{freq_t}$, where $freq_t$ is the number of frames with ground-truth label divided by the total number of frames in the video, and *median_freq* is the median of the computed frequencies.

The mean square error metric is adopted to be the fitting loss $L_{mse}$ of the meta-learner, and the loss function is

$$L_{mse} = \|\mathbf{y}_t - \mathbf{y}\|_2^2 \quad (8)$$

where $\mathbf{y}_t$ is the ground-truth frame-level scores vector, and $\mathbf{y}$ is the meta-learner's output frame-level scores vector.

*F. Key-shot selection*

The Kernel Temporal Segmentation (KTS) algorithm [21] is adopted on the video sequence to calculate the number of shots in the video and their region (i.e., the start and end). And the shot-level importance score is obtained by averaging the final scores output by meta learner in the same shot. Finally, the video summaries are produced under the constraint that the total length of selected shots is no more than 15% of the original video length for a fair comparison with the reference methods, which is implemented by Knapsack algorithm.

III. EXPERIMENTS AND ANALYSIS

*A. Datasets and evaluation metric*

We evaluate our stacking ensemble approach for video summarization (SEVS) on two public benchmark datasets, i.e., TVSum [22] and SumMe [23]. Up to 39 videos from the YouTube dataset and 50 videos from the Open Video Project (OVP) dataset [24] are used to augment the training data. Three settings as suggested in [1] are adopted to evaluate the method as following. (1) Canonical. (2) Augmented. (3) Transfer. Training and testing datasets in all the settings are randomly divided into 5 splits, and the average performance of the 5 runs is achieved. In addition, F-score are used as the metric to assess the similarity between the generated summaries and the ground truth summaries, and diversity score is used to assess the diversity performance of the generated video summaries.

*B. Experiments*

*1) Implementation Details*

The non-maximum suppression threshold is set as 0.5 for the final result presentation. Our model is trained over 300 epochs by using Adam optimizer with a base learning rate of $5\times10^{-5}$ and a weight decay of $10^{-5}$. All the experiments are conducted on a Nvidia GTX 1660Ti GPU and implemented by PyTorch.

## 2) Results and Comparisons

Table I shows the comparison results, where the performance of the other methods is obtained from the corresponding references. It can be seen that the performance of the canonical tests for our model on the two basic datasets is superior to the other methods. Specifically, the F-score on TVSum is increased by at least 1.3%. It is worthy noting that the parameters of our model is fewer than those of most of the models in Table I.

It can also be seen that our method achieves competitive performance compared with the state-of-the-art methods in augmented and transfer settings. Among all the methods, our ensemble model significantly outperforms the two typical methods, i.e., DSNet [14] (a classic temporal interest detection model) and FCSN [2] (a model based on sequence labeling formulation), which validate the effectiveness of our model.

In addition, the diversity performance of our method is assessed on two datasets, and the results are shown in table II. Our method achieves the highest score on SumMe. Compared with DSNet, our method achieves an improvement of 5.5% in the diversity score.

TABLE I
COMPARISONS OF F-SCORE (%) AND PARAMETERS (MILLION) WITH STATE-OF-ART VIDEO SUMMARIZATION METHODS ON THE SUMME AND TVSUM DATASETS UNDER THE CANONICAL (C), AUGMENTED (A) AND TRANSFER (T) SETTINGS, RESPECTIVELY

| Methods | SumMe | | | TVSum | | | Params |
|---|---|---|---|---|---|---|---|
| | C | A | T | C | A | T | |
| vsLSTM [1] | 37.6 | 41.6 | 40.7 | 54.2 | 57.9 | 56.9 | 2.63 |
| dppLSTM [1] | 38.6 | 42.9 | 41.8 | 54.7 | 59.6 | 58.7 | 2.63 |
| FCSN [2] | 48.8 | 50.2 | 45.0 | 58.4 | 59.1 | 57.4 | 116.49 |
| VASNet [3] | 49.7 | 51.1 | - | 61.4 | 62.4 | - | 7.35 |
| SUM-GAN [5] | 41.7 | 43.6 | - | 56.3 | 61.2 | - | 295.86 |
| DR-DSN [6] | 42.1 | 43.9 | 42.6 | 58.1 | 59.8 | 58.9 | 2.63 |
| M-AVS [9] | 44.4 | 46.1 | - | 61.0 | 61.8 | - | - |
| DSNet [14] | 50.2 | 50.7 | 46.5 | 62.1 | 63.9 | 59.4 | 8.53 |
| SABTNet [15] | 50.7 | - | - | 61.0 | - | - | 6.31 |
| SASUM$_{sup}$ [25] | 45.3 | - | - | 58.2 | - | - | 44.07 |
| [26] | 51.1 | 52.1 | 45.4 | 61.0 | 61.5 | 55.1 | - |
| DHAVS [27] | 45.6 | 46.5 | 43.5 | 60.8 | 61.2 | 57.5 | - |
| [28] | 51.7 | 51.0 | 44.1 | 61.5 | 61.2 | 58.9 | - |
| [29] | 51.7 | - | - | 59.6 | - | - | - |
| CSNet$_{sup}$ [30] | 48.6 | 48.7 | 44.1 | 58.5 | 57.1 | 57.4 | - |
| SEVS (Ours) | **51.8** | **51.4** | **46.7** | **62.9** | **63.3** | **59.0** | 4.33 |

TABLE II
THE DIVERSITY SCORE OF GENERATED SUMMARIES ON THE SUMME AND TVSUM DATASETS

| Dataset | dppLSTM | DR-DSN | DSNet | SEVS (Ours) |
|---|---|---|---|---|
| SumMe | 0.591 | 0.594 | 0.642 | **0.676** |
| TVSum | 0.463 | 0.464 | 0.476 | **0.473** |

Notes: The higher the score, the better the diversity of the summary video.

## C. Ablation studies and analysis

Ablation experiments are conducted to verify the effectiveness of our method, and the results are shown in Table III. The experiments are tested separately for the two branches, i.e., temporal interest segments and frame-level probabilities in our model. We also compare different soft decision fusion strategies, i.e., averaging the two scores with the same weight and meta-learning. It can be seen that no matter which soft decision fusion strategy is used, the improvement of the final performance is obvious. Meta-learning strategy achieves higher generalization performance due to the fact that the concurrent learning of segment features and frame-level features makes the model more robust with restraining overfitting.

It can also be seen from Fig. 3 that our method can better fit the ground-truth important scores curve compared with other strategies in the ablation test, which means that our meta learning stacking method could refine the importance scores of each frame under the constraints of temporal interest segments.

The results of the parameter analysis of the NMS threshold on the SumMe and TVSum datasets are shown in Fig. 4. By comparing the F-score and balancing inference time consumption, 0.5 is specified as the final threshold.

TABLE III
F-SCORE RESULTS (%) VIA ABLATION STUDIES ABOUT INTEREST SEGMENTS, FRAME-LEVEL PROBABILITIES, AVERAGING AND META LEARNING STRATEGY

| segments | frame | Avg. | meta | SumMe | | | TVSum | | |
|---|---|---|---|---|---|---|---|---|---|
| | | | | C | A | T | C | A | T |
| ✓ | | | | 50.1 | 49.7 | 44.6 | 62.6 | 62.4 | 58.7 |
| | ✓ | | | 47.0 | 48.6 | 45.9 | 61.6 | 61.9 | 58.5 |
| ✓ | ✓ | ✓ | | 49.8 | 50.6 | 46.5 | 62.8 | 63.0 | 58.2 |
| ✓ | ✓ | | ✓ | **51.8** | **51.4** | **46.7** | **62.9** | **63.3** | **59.0** |

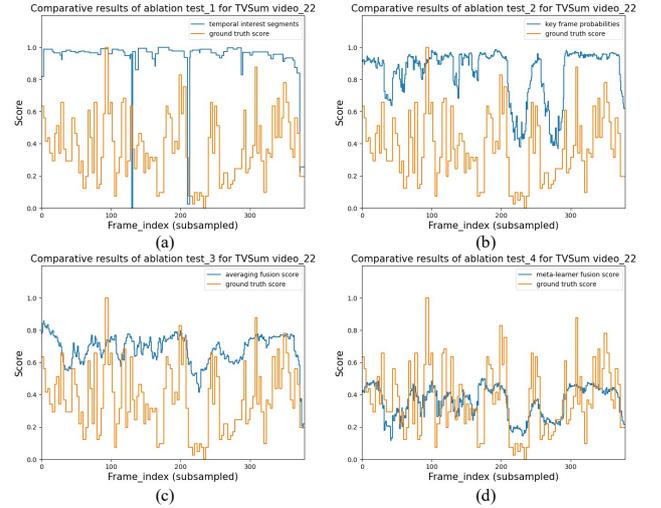

Fig. 3. Comparative step lines via ablation studies about interest segments, frame-level probabilities, averaging and meta learning strategy

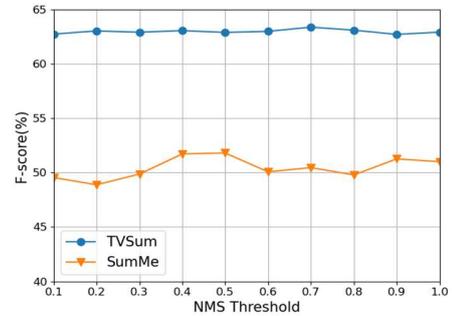

Fig. 4. Parameter analysis of the NMS threshold on the SumMe and TVSum datasets

## IV. CONCLUSIONS

A stacking ensemble approach for video summarization is proposed to exploit the potential complementarity between shot-level and frame-level partition. Comparative experimental results demonstrate the superiority of our model, while a series of ablation experiments show that interest segments scores and frame-level probabilities are potentially correlated, and their fusion can result in more proper decisions than a single method. In the next step, ensemble learning algorithms for video summarization will be further explored.